\documentclass[a4paper, 10pt, DIV=12, headinclude=false, footinclude=false]{scrartcl}
\usepackage[T1]{fontenc}
\usepackage[utf8]{inputenc}
\usepackage[english]{babel}

\makeatletter 
  \newcommand\figcaption{\def\@captype{figure}\caption} 
  \newcommand\tabcaption{\def\@captype{table}\caption} 
\makeatother

\usepackage[pagebackref,breaklinks,colorlinks]{hyperref}

\usepackage{graphicx}

\usepackage{tikz}
\usepackage{comment}
\usepackage{amsmath,amssymb} 
\usepackage{booktabs}
\usepackage{mathtools}
\usepackage{multirow}
\usepackage{color}
\usepackage{amsfonts}
\usepackage{wrapfig}

\usepackage[capitalize]{cleveref}

\usepackage{caption}
\usepackage{wrapfig}
\usepackage{sidecap} 
\usepackage{csquotes}
\usepackage{url}
\usepackage{mathtools}
\usepackage{xfrac}
\usepackage[
   algo2e,
   lined,
   boxed,
  commentsnumbered,
  linesnumbered,
  ruled
 ]{algorithm2e}
\usepackage{xparse}

\usepackage{tikz}
\usetikzlibrary{external}
\usetikzlibrary{arrows,positioning,calc,chains}

\usepackage{pgfplots}
\usepackage{pgfkeys}
\def\addlegendimage{\csname pgfplots@addlegendimage\endcsname}

\usepackage{color}
\usepackage{setspace}
\definecolor{sixclassRdYlBu1}{rgb}{0.84,0.19,0.15}
\definecolor{sixclassRdYlBu2}{rgb}{0.99,0.55,0.35}
\definecolor{sixclassRdYlBu3}{rgb}{1.0,0.88,0.56}
\definecolor{sixclassRdYlBu4}{rgb}{0.88,0.95,0.97}
\definecolor{sixclassRdYlBu5}{rgb}{0.57,0.75,0.86}
\definecolor{sixclassRdYlBu6}{rgb}{0.27,0.46,0.71}

\usepackage{subcaption} 
\DeclarePairedDelimiterX\mengenA[1]{\lbrace}{\rbrace}{#1}
\DeclarePairedDelimiterX\mengenB[2]{\lbrace}{\rbrace}{#1\, \delimsize\vert \, #2}

\makeatletter
\newcommand{\set}[2][\relax]{
	\ifx#1\relax \ensuremath{
		\mengenA*{#2}}
	\else \ensuremath{%
		\mengenB*{#1}{#2}}
	\fi}
\makeatother

\DeclareRobustCommand{\minwidthbox}[2]{%
	\ifmmode
	\expandafter\mathmakebox
	\else
	\expandafter\makebox
	\fi
	[\ifdim#2<\width\width\else#2\fi]{#1}%
}

\DeclareFontFamily{U}{matha}{\hyphenchar\font45}
\DeclareFontShape{U}{matha}{m}{n}{
      <5> <6> <7> <8> <9> <10> gen * matha
      <10.95> matha10 <12> <14.4> <17.28> <20.74> <24.88> matha12
      }{}
\DeclareSymbolFont{matha}{U}{matha}{m}{n}
\DeclareFontSubstitution{U}{matha}{m}{n}

\DeclareFontFamily{U}{mathx}{\hyphenchar\font45}
\DeclareFontShape{U}{mathx}{m}{n}{
      <5> <6> <7> <8> <9> <10>
      <10.95> <12> <14.4> <17.28> <20.74> <24.88>
      mathx10
      }{}
\DeclareSymbolFont{mathx}{U}{mathx}{m}{n}
\DeclareFontSubstitution{U}{mathx}{m}{n}

\DeclareMathDelimiter{\vvvert}{0}{matha}{"7E}{mathx}{"17}

\DeclarePairedDelimiterXPP{\snorm}[1]{}{\lVert}{\rVert}{_{1}}{\ifblank{#1}{\:\cdot\:}{#1}}
\DeclarePairedDelimiterXPP{\anorm}[1]{}{\lVert}{\rVert}{_{a,$\mu$}}{\ifblank{#1}{\:\cdot\:}{#1}}
\DeclarePairedDelimiterXPP{\Snorm}[1]{}{\vvvert}{\vvvert}{_{1}}{\ifblank{#1}{\:\cdot\:}{#1}}
\DeclarePairedDelimiterXPP{\Anorm}[1]{}{\vvvert}{\vvvert}{_{a,$\mu$}}{\ifblank{#1}{\:\cdot\:}{#1}}
\DeclarePairedDelimiterXPP{\SMnorm}[1]{}{\vvvert}{\vvvert}{_{1,$\mu$}}{\ifblank{#1}{\:\cdot\:}{#1}}

\makeatletter
\newcommand{\labeltext}[2]{%
  \@bsphack
  \csname phantomsection\endcsname 
  \def\@currentlabel{#1}{\label{#2}}%
  \@esphack
}
\makeatother

\allowdisplaybreaks

\DeclareOldFontCommand{\rm}{\normalfont\rmfamily}{\mathrm}

\usepackage[pagebackref,breaklinks,colorlinks]{hyperref}
\usepackage{float}

\begin{document}

\title{QS-Craft: Learning to Quantize, Scrabble and Craft for Conditional Human Motion Animation}

\author{Yuxin Hong, ~Xuelin Qian, ~Simian Luo, ~Xiangyang Xue, ~Yanwei Fu  \\[0.1in]
Fudan University}
\date{}

\maketitle

\begin{abstract}
This paper studies the task of conditional Human Motion Animation (cHMA). Given a source image and a driving video, the model should animate the new frame sequence, in which the person in the source image should perform a similar motion as the pose sequence from the driving video. Despite the success of Generative Adversarial Network (GANs) methods in image and video synthesis, it is still very challenging to conduct cHMA due to the difficulty in efficiently utilizing the conditional guided information such as images or poses, and generating images of good visual quality.
To this end, this paper proposes a novel  model of learning to Quantize, Scrabble, and Craft (QS-Craft) for conditional human motion animation. The key novelties come from the newly introduced three key steps:  quantize, scrabble and craft. Particularly, our QS-Craft employs transformer in its structure to utilize the  attention architectures. The guided  information is represented as a pose coordinate sequence extracted from the driving videos. Extensive experiments on human motion datasets validate the efficacy of our model. 
\par\vskip\baselineskip\noindent
\textbf{Keywords}: Human motion animation, transformer, conditional animation, scrabble
\end{abstract}

\section{Introduction}
\label{sec:intro}
The task of 
conditional Human Motion Animation (cHMA)
has attracted increasing attention in the vision community, as it can be utilized in many industrial applications such as computer games,  advertisement, and animation industry ~\cite{cai2018deep,ma2017pose,ma2018disentangled}.
In this work, we aim to solve the cHMA problem by transferring driving videos to animate the humans in the source image. 
Particularly, given a source image and a driving  video,  the  cHMA model  should  animate  the  new  frame  sequence,  in  which  the  person  in  the  source  image  should perform similar motion as pose sequence from the driving video.

\begin{figure}
\begin{centering}
\includegraphics[scale=0.48]{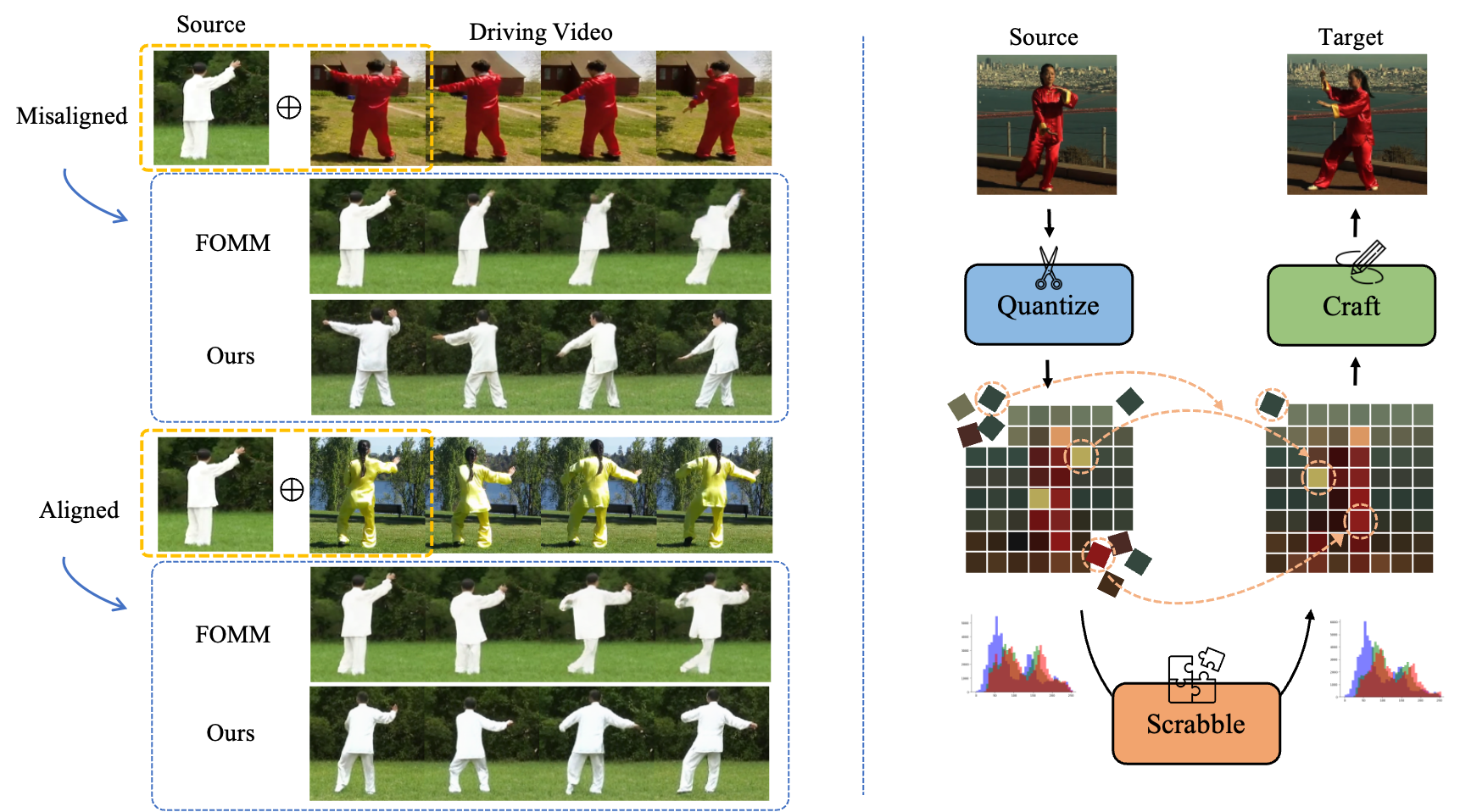}
\vspace{-0.1in}
\caption{Left: Examples of cHMA task. Right: idea illustration of our QS-Craft. Our method learns to quantize  source image, scrabble quantized representation conditioned on driving video, and finally craft the synthesis of target images. Notice that when pose misalignment occurs (left-top), FOMM~\cite{siarohin2019first} fails to generate realistic motion transfer images, while our method accomplishes the task well and preserves finer details. Two histograms (right-bottom) indicate quite similar visual features of the source and target, producing intuition to play the \emph{Scrabble} game.\label{fig:intro}}
\vspace{-0.15in}
\end{centering}
\end{figure}

To tackle this task, 
some classical approaches~\cite{blanz1999morphable,cao2014displaced,thies2016face2face}  are very domain-specific and rely on parametric models to represent humans. 
Generative models, such as Variational Auto-Encoders (VAE)~\cite{kingma2013auto,sonderby2016ladder} and Generative Adversarial Networks (GANs)~\cite{berthelot2017began,mao2017least,qi2020loss,radford2015unsupervised}, have shown good capacity on the image generation task, which could be the alternative ways to address our cHMA task. Unfortunately, these models typically demand expensive training from large collection of data in an \textit{unconditioned} manner.
On the other hand, flow-based approaches~\cite{siarohin2019first,siarohin2019animating,siarohin2021motion} have received more and more attention due to its remarkable performance. The key idea of these methods is to learn motion flows between the source image and the driving frames, such that the target image features can be synthesized by wrapping. For cHMA, authours~\cite{siarohin2019animating,siarohin2021motion,siarohin2019first} proposed to use keypoints to produce flow maps, but these works additionally require modules to map the flow from sparse to dense.  This is not only  difficult to be learned but also may introduce  lots of artifacts.
For example, the misalignment of human motion in \cite{siarohin2019first} in the source image and the first driving video frame may lead to  an unreasonable animated result as shown in Fig.~\ref{fig:intro}(left column). 


In this paper, we advocate the flow-based way yet from a different and novel perspective. As illustrated in Fig.~\ref{fig:intro}, we propose to generate the target image by only rearranging the order of patches in the source image, on account of the observation that the source and the target images have very similar distribution. This is similar to a game of `Scrabble' using the same letters
to spell different words but following different rules. So, the key question is \textit{how to learn the rule of Scrabble?} We argue that it should satisfy three requirements: (i) rules should be objective and independent of input; (ii) the order of patches should be reasonable, including the texture of background and the identity of foreground; (iii) obviously it should be conditioned on the driving videos. To this end, we present a novel paradigm of learning to \textbf{Q}uantize, \textbf{S}crabble and \textbf{C}raft for conditional human motion animation (cHMA), thus dubbed  QS-Craft. 

Formally, 
given the input source image and the driving video, our QS-Craft model should have the following three key steps, as illustrated in Fig.~\ref{fig:intro} (Rigth column).
 (1) \emph{Quantize}. Train an encoder to get the latent feature of the source image and then quantize it to produce 
 a bag of discrete feature pixels by referring to the global codebook, which is  trained in an end-to-end manner. (2) \emph{Scrabble}. Conditioned on motion information from  driving video, our QS-Craft trains a decoder-only transformer to rearrange the feature pixels computed in (1) to produce the corresponding feature pixels for the targets. By virtue of such a way, we can effectively exploit the distribution of visual features, to help synthesize the latent features of the target images.
 (3) \emph{Craft\footnote{The name of Craft is inspired by the game of Minecraft.}}. Finally, the reordered feature pixels
 will be passed to a decoder layer to produce the animated target images. 
 Extensive experiments on benchmarks validate the efficacy of our model, and the quantitative and qualitative results show that our method outperforms the state-of-the-art competitors.

\noindent \textbf{Contributions}.
The main contribution of this paper is to propose a novel paradigm for  the conditional human motion animation: learning to Quantize, Scrabble and Craft  (QS-Craft). 
The {quantize} step is first introduced to encode the input source image and driving videos into a discrete representation. Next, we employ the transformer to learn to \textit{scrabble} by mixing up pixel distributions and exploiting the known distribution of the source image to fit those in driving videos. The reordered feature pixels are decoded to \textit{craft} the human motion animation. Based on this novel model framework, we can not only address motion transfer when pose misalignment occurs (shown in Fig.~\ref{fig:intro}) but achieve better quantitative scores in video reconstruction compared to other state-of-the-art methods.

\section{Related Works}
\noindent  \textbf{Image Animation. }Image animation and video re-targeting have drawn attention of computer vision researchers in recent years. Some of the previous approaches~\cite{cao2014displaced,thies2016face2face,blanz1999morphable} are heavily domain-specific, which means they can only tackle animation of human faces, human silhouettes, etc. When transferred to other situations, these methods might easily fail and are ineffective. One given driving video and one source image attempt to guide models generating another video in the domain of the source image with corresponding motion information in driving video. Recycle-GAN~\cite{bansal2018recycle} incorporates Spatio-temporal cues into a conditional GAN architecture to translate from one domain to another. However, it has to be trained with a strong prior of specific domains and cannot be generalized to even another individual. Similarly, the model in \cite{wang2018video} aiming to address the motion transfer task is also exposed to the same problem. Compared to these works, our method does not need any prior during the training phase and can be domain-agnostic in the inference time.
X2Face~\cite{wiles2018x2face}  employs reference poses to warp a given image and obtain the final transferred video. It does not require priors but is better to face animations than cHMA. Siarohin \emph{et al.} introduced Monkey-Net~\cite{siarohin2019animating}. This network can indeed animate arbitrary objects guided by detected key points, but it is only effective on lower resolution images. The authors in \cite{siarohin2019first} and \cite{siarohin2021motion} proposed an affine transformation and a PCA-based motion estimation respectively to transfer driving motions into the target images with higher resolution. Note that both of them are based on warping motion flow in order to encode images in feature space then decoding to get the targets. It suggests that the latent feature in target images has to be encoded from scratch whilst our proposed model can  rearrange the quantized feature sequences conditioned on driving videos for the motion transfer task. This discretization method ensures better quality of our generated images. Moreover, the main structure of our model is transformer, which is quite different from other methods mentioned above. And both the quantitative and qualitative results show the efficacy of this prominent block.

\noindent \textbf{Visual Transformer. }Since the introduction of the self-attention module~\cite{vaswani2017attention}, many transformer-based methods have achieved impressive performance in numerous Natural Language Processing (NLP) and vision tasks~ \cite{radford2018improving,devlin2018bert,dosovitskiy2020image,carion2020end,liu2021swin}. Autoregressively generating pixels  by transformer can only be applied to low-resolution images due to costly computation and huge memory footprint~\cite{chen2020generative,kumar2021colorization}. Many recent works, such as dVAE~\cite{ramesh2021zero} and VQ-VAE~\cite{oord2017neural}, attempt to model discrete representation of images to reduce the total sequence length in order to break through this bottleneck. Besides, VQ-GAN~\cite{esser2021taming} added Generative Adversarial Networks (GANs) into VQ-VAE to improve training efficiency. Both VQ-VAE and VQ-GAN quantize an image according to a global codebook. In the second stage of \cite{esser2021taming}, its transformer module has to look up in the whole codebook to generate latent features of targets, which may lead to accumulated errors and unreasonable generated results. 
Hence constructing a local dynamic codebook for each conditions-target pair can enhance the searching ability of the transformer; and thus the target images generated by the decoder will be more realistic. To achieve this goal, we propose to quantize one image twice to rearrange conditions for the target one. And this scrabbling can also speed up convergence of the model.

\section{Methodology}

The purpose of this paper is to animate the human in a source image $x_{s} \in \mathbb{R}^{H\times W \times 3}$ 
conditioned on the motion in a driving video. For the target image $x_{t} \in \mathbb{R}^{H\times W \times 3}$, it should  be semantically consistent with the source image, including the information of background and the content of the human texture. More importantly, it demands natural and realistic motion changes, which are guided by a series of driving frames (\textit{i.e.}, the condition $c$). Following our inspiration, we assume that the target image $x_{t}$ can be generated by reordering the latent features $z_{s}$ of the source image in spatial dimensions, which can be expressed as,
\begin{equation}
    x_{t} = G\left( \mathcal{T}\left(z_{s},c\right)\right) = G\left( \mathcal{T}\left(E\left(x_{s}\right),c\right)\right)
    \label{eq:overall}
\end{equation}
\noindent where $E\left(\cdot\right)$ and $G\left(\cdot\right)$ denote the encoder and the decoder layer; $\mathcal{T}\left(\cdot\right)$ means the operation of `scrabble', \textit{i.e.}, putting latent features $z_{s}$ together following a specific rule.

\begin{figure}[tbp]
\centering
\includegraphics[scale=0.46]{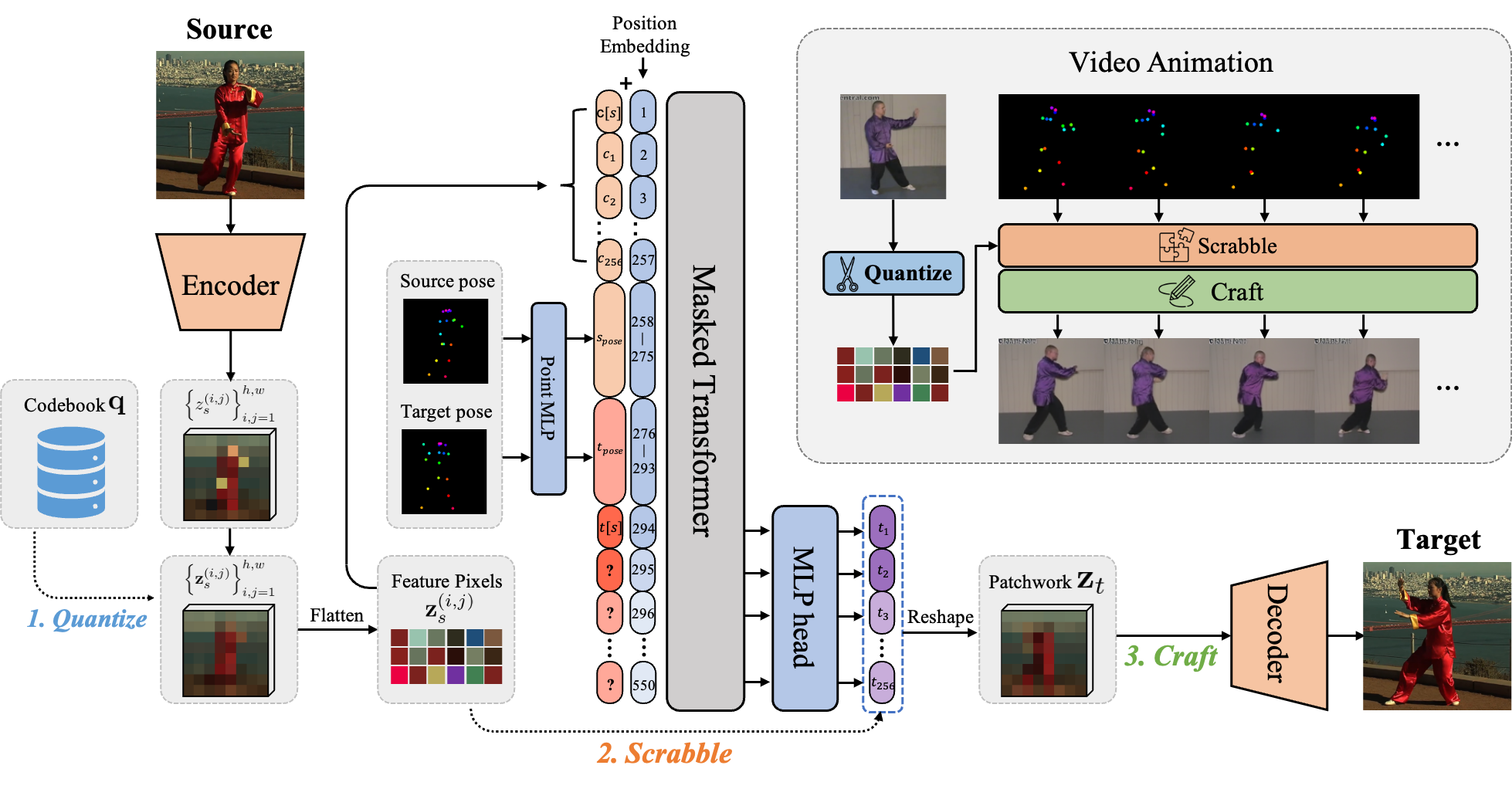}
   \caption{The overall pipeline of QS-Craft. We first train an encoder-decoder based framework to quantize both source and target images. Then, a conditional transformer is designed to learn the scrabble game under the given driving videos. Finally, the decoder crafts the reordered features as output. \label{fig:pipeline}}
   \vspace{-0.1in}
\end{figure}

The overall of our framework is illustrated in Fig.~\ref{fig:pipeline}, which is composed of three steps, \textit{Quantize}, \textit{Scrabble} and \textit{Craft}. First, we perform the vector quantization mechanism~\cite{oord2017neural} on the encoded latent features $z_{s}$ of the source image, resulting in a bag of feature pixels $\left\{\mathbf{z}_{s}^{\left(i,j\right)}\right\}_{i,j=1}^{h,w}$, where $h$ and $w$ indicate the height and width of the latent features, respectively. Then, we play a Scrabble game by selecting pixels from the bag and mixing them together, $\hat{z}_{t} = \mathcal{T}\left(\mathbf{z}_{s},c \right)$. This process must be meaningful and also be conditioned on the motions from driving videos. Finally, the patchwork $\hat{z}_{t}$ is fed to the decoder to craft the image, which is realistic  with natural motion changes. 

Considering that $\mathcal{T}\left(\cdot\right)$ is operated on index-based representation, we divide the training procedure of our model into two stages, \textit{image reconstruction with scrabble} and \textit{learning scrabble rules with transformer}. For the rest of this section, we will follow these two training stages to elaborate our framework in Sec.~\ref{sec:stage1} and Sec.~\ref{sec:stage2}, respectively. 



\subsection{Image Reconstruction with Scrabble \label{sec:stage1}}

In the first stage, we aim to teach the decoder $G$ to generate a natural target image $x_{t}$ from a known patchwork $\hat{z}_{t}$. We first utilize the encoder $E$ to extract latent features from both source and target images, denoted as $z_{s}$ and $z_{t}$. Considering the correlations between each feature pixel in $z_{s}$, we adopt one more step of vector quantization before fitting them together. Specifically, a learnable codebook $ \mathbf{q} \in \mathbb{R}^{m \times c_{q}}$ is introduced, where $m$ and $c_{q}$ are the number and dimension of codes, respectively. For each pixel $\left\{z_{s}^{\left(i,j\right)}\right\}_{i,j=1}^{h,w}$, we do quantization $\mathcal{Q}\left(\cdot\right)$ by replacing it with the closest codebook entry $q_{k}$ in $\mathbf{q}$ \footnote{We refer the readers to \cite{esser2021taming} for details of the codebook learning.},

\begin{equation}
   \mathbf{z}_{s} = \mathcal{Q}\left(z_{s}\right) := \arg\min_{q_{k}\in \mathbf{q}} \lVert z_{s}^{\left(i,j\right)} - q_{k} \rVert 
   \label{eq:vqvae}
\end{equation}

\noindent where $\mathbf{z}_{s}  \in \mathbb{R}^{h\times w \times c_{q}}$. After that, we obtain a bag of quantized feature pixels $\left\{\mathbf{z}_{s}^{\left(i,j\right)}\right\}_{i,j=1}^{h,w}$. 

In order to reconstruct the target image $x_{t}$, we utilize the latent feature $z_{t}$ as a reference to play scrabble with pixels (words) in $\mathbf{z}_{s}$. More concretely, we compute the distance between feature pixels in $z_{t}$ and $\mathbf{z}_{s}$, and select the closest ones to fit a patchwork $\hat{z_{t}} \in \mathbb{R}^{h\times w \times c_{q}}$. The formulation can be written as, 

\begin{equation}
   x_{t} = G\left( \hat{z_{t}} \right),~~~~ \hat{z_{t}} := \arg\min_{z_{k}\in \mathbf{z}_{s}} \lVert z_{t}^{\left(i,j\right)} - z_{k} \rVert
   \label{eq:reconstruct}
\end{equation}

To encourage the fidelity of the synthesized images, we reverse the generation flow in \cref{eq:reconstruct}, that is, using a bag of quantized target feature pixels $\left\{\mathbf{z}_{t}^{\left(i,j\right)}\right\}_{i,j=1}^{h,w}$ to fit a patchwork of the source $\hat{z_{s}}$. Furthermore, we also introduce a perceptual loss \cite{xu2020pose,tulyakov2018mocogan} and a Discriminator $D$ \cite{yang2018pose,ma2017pose} to highly maintain the perceptual quality during training. 

\begin{figure}[tbp]
\begin{centering}
\includegraphics[scale=0.4]{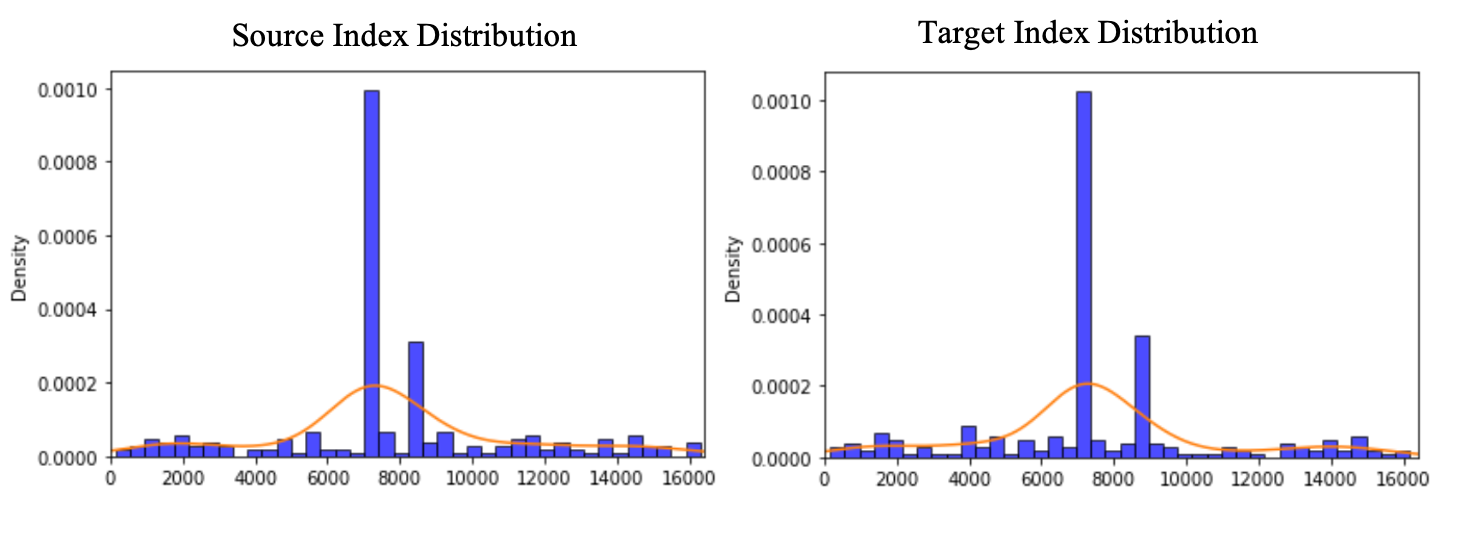}
\vspace{-0.15in}
\caption{Index distribution of a random sample pair $\mathbf{z}_s$ and $\mathbf{z}_t$.  \label{fig:discussion}}
\end{centering}
\vspace{-0.15in}
\end{figure}

\noindent \textbf{Remark.}
Different from \cref{eq:vqvae}, \cref{eq:reconstruct} is much more important here. A straightforward explanation of \cref{eq:reconstruct} is to narrow down the search space in \cref{eq:vqvae}. Experiments in Sec.~\ref{subsec:exp-auantitative} empirically show that this constraint will not degrade the expressive power of the representation $\hat{z_{s}}$ or $\hat{z_{t}}$. Even better, it is efficient and keeps more fidelity in background and foreground. Furthermore, to simply validate our explanation, we randomly choose a sample pair and plot their index distribution in Fig.~\ref{fig:discussion}. As we can see, the index distribution of $\mathbf{z}_s$ and $\mathbf{z}_t$ are quite similar; therefore, by scrabbling the indices in the source, we are able to reconstruct the target image.

\subsection{Learning Scrabble Rules with Transformer}
\label{sec:stage2}

Through the process described in \cref{sec:stage1}, the Decoder $G$ now can synthesize realistic images according to the given patchwork. Nevertheless, the patchwork is generated with the reference of target image, which is not available during inference. To this end, we propose  learning  the rule of Scrabble, such that the desired patchwork $\hat{z}_{t}$ can be successfully produced \textit{only with} a bag of source feature pixels $\mathbf{z}_{s}$ and the condition motion information $c$. 

Here, the condition $c$ refers to the information of pose coordinates. We claim that directly utilizing RGB images as condition may cause the information leakage~\cite{iLAT2021}  when both the source and driving frames are from the same video. Empirical results \cref{subsec:ablation_pose} suggest that pose coordinates can not only provide effective and straightforward condition, but also save lots of computational costs compared with using pose image \cite{esser2021taming}. 
More precisely, an off-the-shelf open pose detector is applied to detect and localize $n$ keypoints of an object. Each point is represented by $x$-$y$ coordinates. If a point is occluded or undetectable, we set $x=y=-1$ for indication. Then, we attach three fully-connected layers with ReLU activation to encode the 2D pose coordinates $p \in \mathbb{R}^{n\times 2}$, which can be expressed as,

\begin{equation}
   c = \mathcal{F}\left(p\right) \in \mathbb{R}^{n\times n_{c}}
   \label{eq:condition}
\end{equation}
\noindent where $\mathcal{F}\left(\cdot\right)$ denotes the stacked fully-connected layers, $n_{c}$ means the output dimension. We flatten the pose condition $c$ as a sequence $\left\{c_{i}\right\}_{i=1}^{n}$, which is further incorporated with our transformer.  

Speaking of the transformer, we adopt a decoder-only structure and first convert the feature representation of $\mathbf{z}_{s}\in \mathbb{R}^{h\times w \times c_{q}}$ and $\hat{z}_{t}\in \mathbb{R}^{h\times w \times c_{q}}$ to the sequence of index representation, \textit{i.e.}, 
$\left\{s_{i}\right\}_{i=1}^{l}$ and $\left\{t_{i}\right\}_{i=1}^{l}$, where its value ranges from $0$ to $m-1$, indicating the index of each feature pixel in the codebook $\mathbf{q}$, and $l=h\times w$. Two embedding layers with the dimension of $n_c$ are subsequently followed for $\left\{s_{i}\right\}$ and $\left\{t_{i}\right\}$, respectively \footnote{We reuse the symbols of $\left\{s_{i}\right\}$ and $\left\{t_{i}\right\}$ after embedding for simplicity.}. During training phase, we concatenate the source embedding $\left\{s_{i}\right\}$ and the condition sequence $\left\{c_{i}\right\}$ as input. 
Note that the previous ground-truth target embedding $\left\{t_i\right\}_{i=1}^{j-1}$ are progressively attached to the input to predict the likelihood of the next target index $t_{j}$. The overall formulation can be written as,

\begin{equation}
   p\left( t | s,c\right) = \prod_{j=1}^{l}p\left( t_{j} | s,c,t_{\left[s\right]},t_{<j} \right)
   \label{eq:autoregress}
\end{equation}
\noindent where $t_{\left[s\right]} \in \mathbb{R}^{1 \times c_{q}}$ denotes a learnable start token, which is added after the condition sequence for the case of predicting the first target index $t_{1}$, as shown in \cref{fig:pipeline}. For inference, we use the previously predicted index to replace the ground-truth ones. With regard to the multi-head self-attention in the transformer, we design a new attention mask $\mathbf{M}$ with four sub-masks,

\begin{equation}
   \mathbf{M} = 
   \begin{bmatrix}
   \mathbf{A} & \mathbf{B} \\
   \mathbf{C} & \mathbf{D}
   \end{bmatrix} = 
   \begin{bmatrix}
   \mathbf{1} & \mathbf{0} \\
   \mathbf{1} & \mathbf{M}_{tril}
   \end{bmatrix}
   \label{eq:mask}
\end{equation}
 
\noindent where $\mathbf{A} \in \mathbb{R}^{\left(l+n\right)\times \left(l+n\right)}$ and $\mathbf{C} \in \mathbb{R}^{l\times \left(l+n\right)}$ are all-ones matrices, designed to learn the relationship among the source embedding $\left\{s_{i}\right\}$, the conditional information $\left\{c_{i}\right\}$ and the target embedding $\left\{t_{i}\right\}$; $\mathbf{M}_{tril} \in \mathbb{R}^{l\times l}$ means a standard lower triangular matrix, so that the next target index can be deduced from the previous known information.

Recall the key idea of `Scrabble' in the transformer, a patchwork $\hat{z}_{t}$ is acquired \textit{only with} feature pixels from $\mathbf{z}_{s}$, thus we apply a mask constraint to the output of index probabilities. Specifically, denote the output of the $j$-th target index from transformer as $v_{j} \in \mathbb{R}^{m}$, we mask out some elements if their indices are out of the bag of source index $\left\{s_{i}\right\}_{i=1}^{l}$,

\begin{equation}
   v_{j}^{k} = -\inf~~if~~k \notin \left\{s_{i}\right\}_{i=1}^{l}
   \label{eq:output}
\end{equation}

\noindent where $k \in \left[0, m-1\right]$ and $v_{j}^{k}$ indicates the $k$-th element in $v_{j}$. We feed masked $v_{j}$ into $Softmax$ operation to obtain final probability of the $j$-th target index. 
Furthermore, in order to improve the fidelity and consistency of the synthesized images, especially for the foreground objects, we encourage the model to learn more correlations between target indices by re-weighting the loss of each predicted index. Since we intuitively strengthen the learning of the foreground area, we call this strategy as RoI (regions of interest) weight.

\section{Experiments}

\noindent \textbf{Datasets.} We evaluate our model on two widely-used benchmarks. (1) Tai-Chi-HD, collected from YouTube following \cite{tulyakov2018mocogan}, is a dataset of human bodies performing Tai Chi actions, consisting of 252 videos for training and 28 videos for testing. 
	Following MRAA preprocessing method, we finally obtain $3,049$ and 285 video chunks for training and testing.
	All video frames are resized to 256 $\times$ 256.
(2) Penn Action (PA) dataset~\cite{6751390} contains $2,326$ video sequences of 15 action classes. All video frames are resized to 256 $\times$ 256 after preprocessing. 

\noindent \textbf{Metrics.} In our experiments, we use the following metrics to provide an in-depth comparisons with other competitors. 
 (1)
  \textit{Average Keypoint Distance (AKD)}, which means the average distance between the detected keypoints of the ground truth image sequences and the generated ones. For both two datasets, we employ the human-pose estimator in \cite{cao2017realtime}
  (2) \textit{Missing Keypoint Rate (MKR)}, Another metric evaluating the difference between the poses in the real images and the reconstructed, is the proportion of keypoints detected in  ground truth but missing in the reconstruction.
    (3) \textit{Fr\'echet Inception Distance (FID)}~\cite{heusel2017gans}, which measures the quality of generated images. 
    In this paper, we concentrate more on the foreground area (\emph{i.e.}, human bodies) to evaluate the fidelity and consistency.

\begin{table}[tbp] 
\small
\centering
\caption{Pose-guided Generation Results on PA dataset. \label{tab:pa_pose-guided}}
\setlength{\tabcolsep}{8mm}{
\begin{tabular}{l|ccc}
\hline
\multicolumn{1}{c|}{\multirow{2}{*}{Methods}}	   & \multicolumn{3}{c}{Penn Action}         \\ \cline{2-4}
			   & AKD $\downarrow$& MKR  $\downarrow$  &FID   $\downarrow$       \\
		\hline
		PG2\cite{ma2017pose}      & 20.5767   & 0.2793      & 78.6146                        \\
		PN-GAN\cite{qian2018pose} & 19.6366   & 0.1666      & 47.0964                        \\
		PATN\cite{zhu2019progressive}  & 19.2875  & 0.2676  & 51.5828                        \\
		MR-Net\cite{xu2020pose}   & \textbf{13.6633} & 0.1688 	& 58.7962 \\ 	\hline
		Ours                      & 18.0123   & \textbf{0.1183} & \textbf{32.5837}                        \\ \hline
		\end{tabular}}
\end{table}

\begin{figure}[tb]
\centering
\begin{tabular}{c}
\includegraphics[scale=0.4]{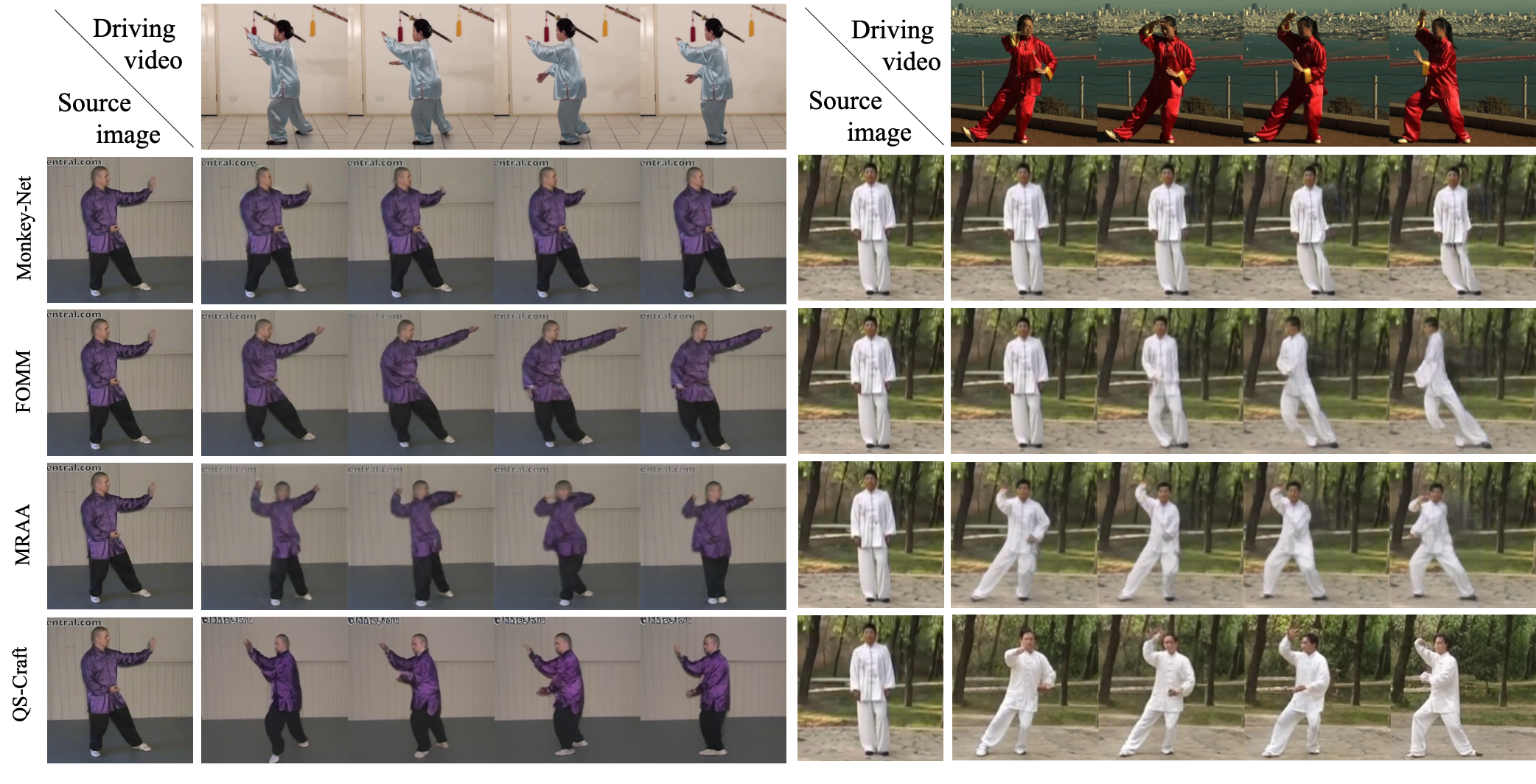}
\tabularnewline
\end{tabular}
  \vspace{-0.15in}
   \caption{Visualization of image animation on Tai-Chi-HD. Given the source image and driving videos, our animated images are better than those from competitors. \label{fig:animation1}}
    \vspace{-0.15in}
\end{figure}

\noindent \textbf{Implementation Details.}
On all datasets, we train our model in two stages. 
(1) for the reconstruction training, our generative model is finetuned on ImageNet datasets~\cite{esser2021taming}. In each training iteration, we randomly select $2$ frames from the same video for training. Adam optimizer is applied with $\beta_1=0.5, \beta_2=0.9$, batch size $2$ and the initial learning rate of $5e$-$5$. We drop the learning rate by half for every $70$K steps. The total training iteration for the first stage is about $210$K.
(2) For the transformer training, we use Adam optimizer with $\beta_1=0.9, \beta_2=0.95$ and batch size $12$. 
We do not change the initial learning rate but add the warmup strategy within the first $10$K steps. The learning rate is linearly decayed to 0 gradually. We train the second stage $280$K for Tai-Chi-HD and $270$K for PA.

\noindent \textbf{Competitors.}
Several related models are listed as competitors. For Tai-Chi-HD dataset, we compare our QS-Craft with three state-of-the-art models for animating, namely, Monkey-Net~\cite{siarohin2019animating}, FOMM~\cite{siarohin2019first} and MRAA~\cite{siarohin2021motion} and report both qualitative and quantitative results. For PA dataset, as for its complexity and variety of human motions, it is not suitable for the animation task whereas tends to be a dataset of pose-guiding. Hence, we compare with four pose-guided generative models, including PG$^{2}$~\cite{ma2017pose}, PATN~\cite{zhu2019progressive}, PN-GAN~\cite{qian2018pose} and MR-Net~\cite{xu2020pose}. 

\subsection{Qualitative Results}
\label{subsec:exp-qualitative}

Note that the objective metrics used in our cMHA task can only reflect the general quality of synthesized images, while it is difficult for these metrics to directly evaluate whether the synthesized images are mimicking the human motion sequences from the driving video. In that case, the qualitative evaluations are much more important in cMHA task to directly reflect whether our model works well. 


\begin{table}[tbp]
	\begin{minipage}[t]{0.48\linewidth}
		\centering
		\small{
		\caption{Video Reconstruction Results on Tai-Chi-HD. \label{tab:taichi_reconstruction}}
		\setlength{\tabcolsep}{0.5mm}{
		\begin{tabular}{l|ccc}
        \hline
        \multicolumn{1}{c|}{\multirow{2}{*}{Methods}} & \multicolumn{3}{c}{Tai-Chi-HD}   \\ \cline{2-4}
         & AKD $\downarrow$     & MKR $\downarrow$   & FID $\downarrow$ \\
         \hline
        Monkey-Net~\cite{siarohin2019animating} & 13.77 & 0.061  & 32.223       \\
        FOMM~\cite{siarohin2019first}           & 7.02  & 0.031  & 25.910   \\
        MRAA~\cite{siarohin2021motion}          & 5.73  & 0.025  & 35.794   \\
        \hline
        Ours     & \textbf{4.61}					& \textbf{0.017} & \textbf{25.064} \\ \hline
	\end{tabular}}}
	\end{minipage}
	\hspace{0.01in}
	\begin{minipage}[t]{0.5\linewidth}
		\centering
		\small{
		\caption{Quantitative ablation study for pose-guided generation on PA dataset.
 \label{table:pa_ablation}}
	    \small{
		\setlength{\tabcolsep}{0.6mm}{
		\begin{tabular}{l|ccc}
\hline
\multicolumn{1}{c|}{\multirow{2}{*}{Methods}}    & \multicolumn{3}{c}{Penn Action}  \\ \cline{2-4} 
	   & AKD $\downarrow$& MKR  $\downarrow$  &FID   $\downarrow$                 \\ 
	   \hline
\emph{w/o} \emph{Scrabble}       & 31.878 &  0.326 & 42.721 \\
\emph{w/o} RoI weight                 & 17.819 &  \textbf{0.121} & 33.066 \\  \hline
Full Model                      & \textbf{16.358} & \textbf{0.121} & \textbf{30.136}                        \\ \hline
\end{tabular}}}}
	\end{minipage}
\end{table}

\begin{figure}[tbp]
\centering
\begin{tabular}{c}
\includegraphics[scale=0.4]{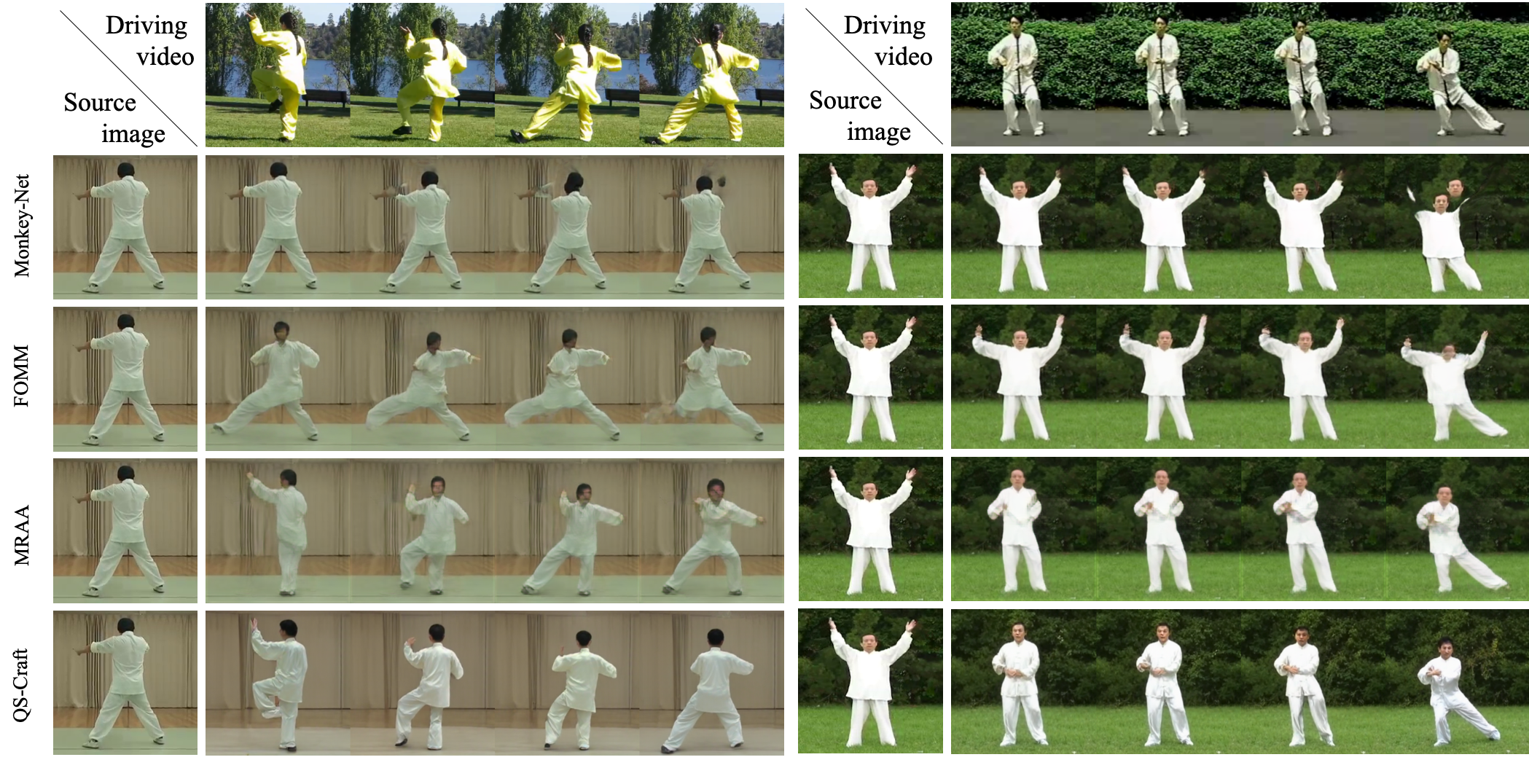}
\tabularnewline
\end{tabular}
\vspace{-0.15in}
   \caption{More animation results on Tai-Chi-HD. Best viewed in color and zoom in.
   \label{fig:animation2}
}
\end{figure}

\noindent \textbf{Human Animation.} 
\Cref{fig:animation1,fig:animation2} show animation results on Tai-Chi-HD dataset. It can be noticed that our method can generate more realistic images with accurate motion according to driving videos on the four randomly selected samples. As these competitors all highly rely on flow information and warping operations, once the source image is not aligned with the first frame of the given driving video, it will introduce either wrong poses or twisted human bodies in the following generated animation results. For example, in the left subfigure of \cref{fig:animation1}, results of Monkey-Net in the first row indicate the failure of motion animation: it only copies the source image. In contrast, our QS-Craft introduces the \textit{Scrabble} step to rearrange the discrete pixel representations from the source image conditioned on the driving motion, thus facilitating the large human poses between the source and driving videos.
Moreover, FOMM can sometimes capture action information in the driving but collapses in most cases. For MRAA, it can indeed generate the animated human bodies with roughly correct poses but details such as ``raising left foot'' are missing. 
It should be emphasized that FOMM tends to fail in the case of misalignment between human motion in the source image and the first frame of driving video, which has been discussed in the limitation of its original paper. As for MRAA, it alleviates the misalignment problem to a certain extent but at the cost of losing a lot of details.
Compared to these methods, our proposed QS-Craft can transfer source images accurately as well as retain most important details, thanks to the scrabble step which can well mirror and exploit the distributions of source images.  
It is apparent that qualitative results in the remaining three subfigures are similar to the first one.

\begin{figure}[tbp]
\centering
\begin{tabular}{c}
\includegraphics[scale=0.5]{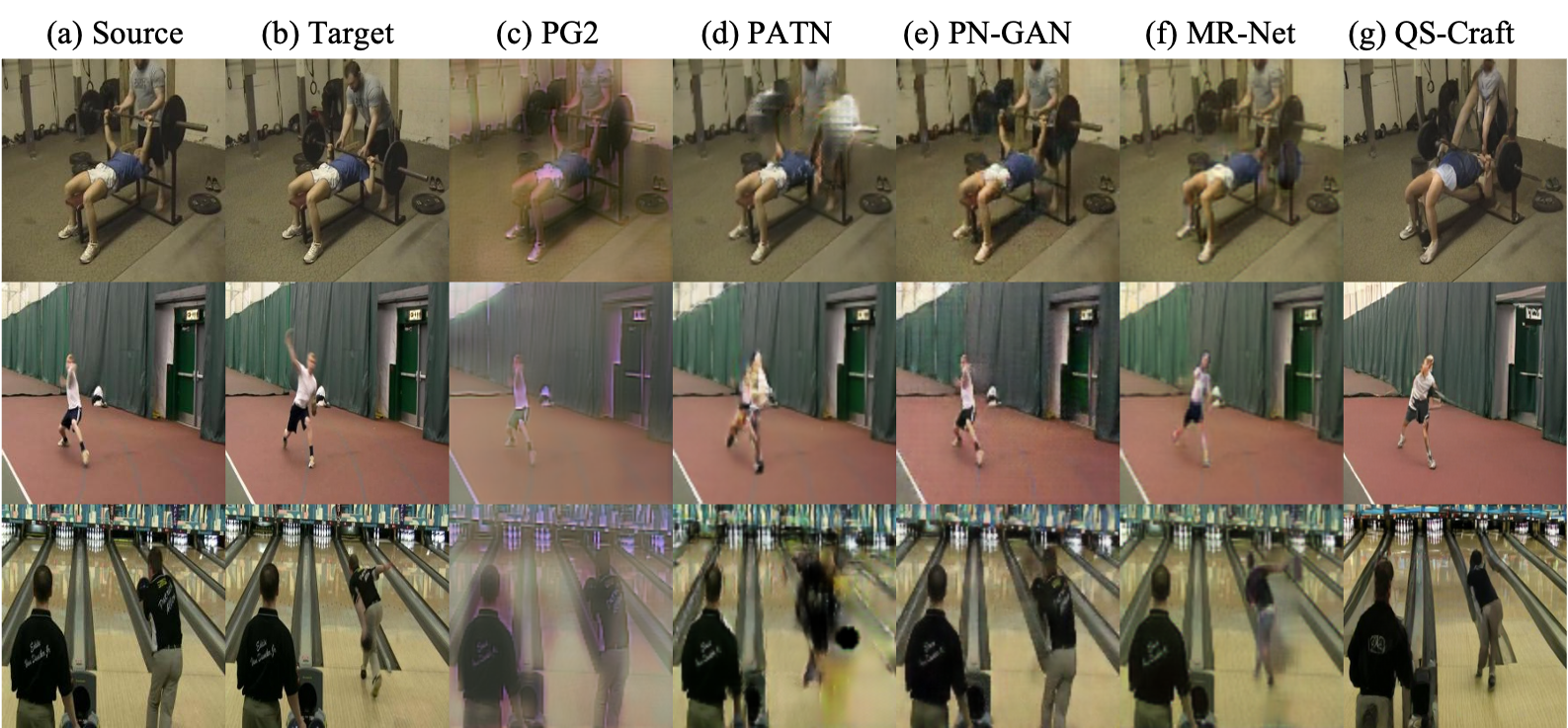}
\end{tabular}
 \vspace{-0.15in}
 \caption{Visualization of pose-guided generation on PA dataset. 
 \label{fig:pa_poseguided}}
\end{figure}

\begin{figure}[tp]
\centering
\begin{tabular}{c}
\includegraphics[width=0.9\textwidth]{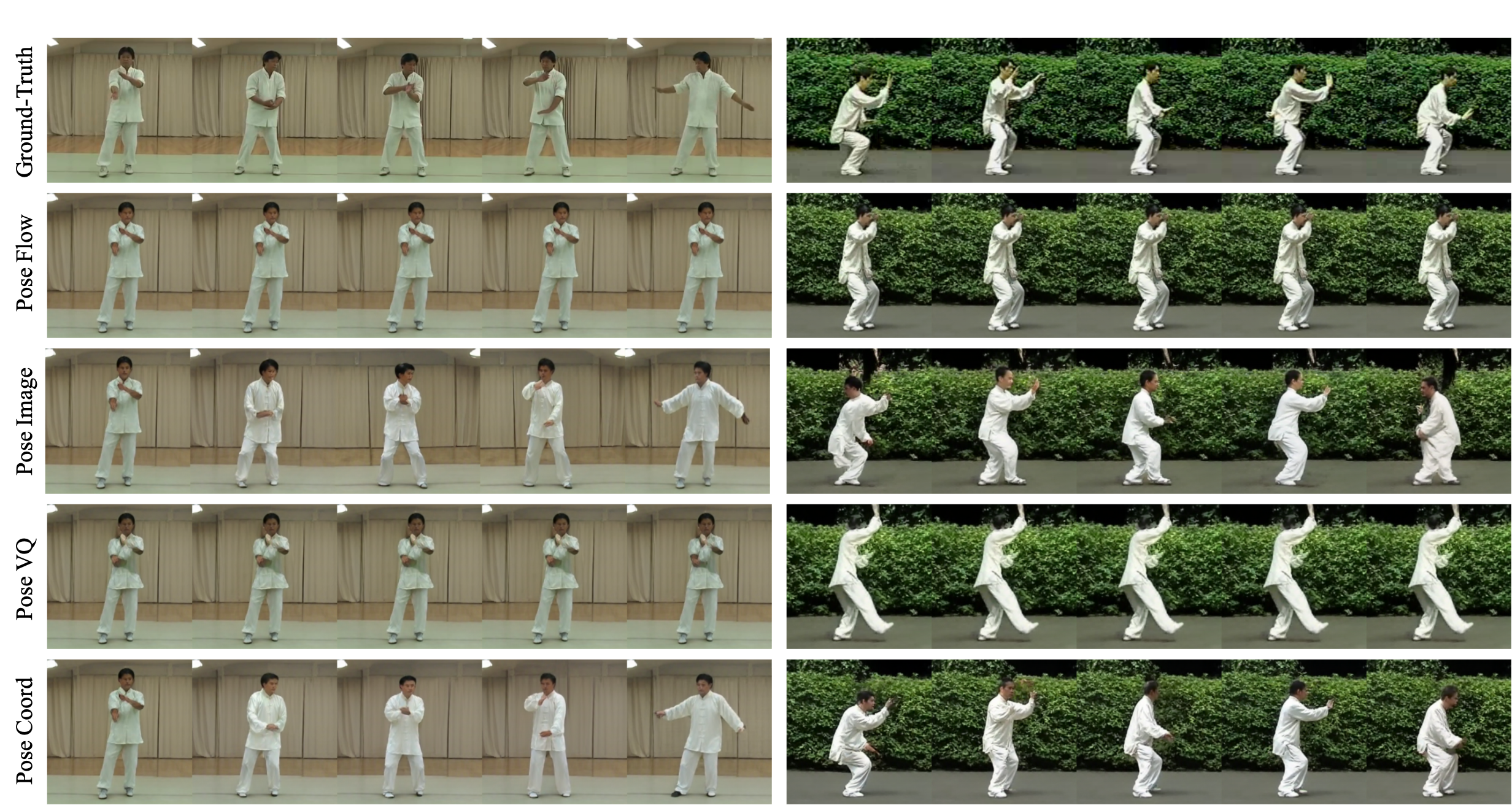}
\end{tabular}
 \vspace{-0.15in}
 \caption{Qualitative comparisons of image reconstruction with different conditions.
 \label{fig:ablation}} 
\end{figure}

\noindent \textbf{Pose Guidance. } We further provide results on PA dataset, which is a more challenging and large dataset.
The qualitative results are present in Fig.~\ref{fig:pa_poseguided}.  We compare the competitors of PG$^{2}$, PATN, PN-GAN and MR-Net. 
Particularly, we show that PG$^{2}$ tends to generate blur results in most cases. Besides, PN-GAN is inclined to copy the source image as target ones. Human body parts in results of PATN are twisted, resulting in unreasonable artifacts.
For MR-Net, the area around the human is blurred, making the synthesized images unrealistic. 
Compared to these models, our QS-Craft demonstrates its efficiency in generating the accurate human poses guided by conditions.

\begin{table}[h]
	\begin{minipage}[t]{0.53\linewidth}
		\centering
		\small{
		\tabcaption{Quantitative comparisons of image reconstruction with different conditions. \label{tab:ablation}}
		\setlength{\tabcolsep}{1.5mm}{
		\begin{tabular}{l|cccc}
 \hline
 \multicolumn{1}{c|}{\multirow{2}{*}{Methods}} & \multicolumn{3}{c}{Tai-Chi-HD} \\ \cline{2-4}
       & AKD $\downarrow$   & MKR $\downarrow$   & FID $\downarrow$    \\ \hline
 Pose Flow           & 17.630   & 0.084      & 56.447         \\
 Pose Image             & 11.715   & 0.033      & 35.526      \\
 Pose VQ    &     18.325      &      0.057        &  53.283               \\
 Pose Coord      & \textbf{5.962}   & \textbf{0.023}      & \textbf{27.784}        \\ \hline
    \end{tabular}}}
	\end{minipage}
	\hspace{0.08in}
	\begin{minipage}[t]{0.42\linewidth}
		\centering
		\small{
		\caption{User study results on Tai-Chi-HD: user preferences
in favour of our approach. \label{tab:taichi_user_study}}
	    \small{
		\setlength{\tabcolsep}{1.8mm}{
		\begin{tabular}{l|ccc}
 \hline
\multicolumn{1}{c|}{\multirow{2}{*}{Methods}} & \multicolumn{2}{c}{Tai-Chi-HD}   \\ \cline{2-3}
 & Test 1     & Test 2  \\ \hline
FOMM~\cite{siarohin2019first}    & 19.87\%    & 25.29\%       \\
Ours     & 80.13\%         & 74.71\%   \\ \hline
 \end{tabular} }}}
	\end{minipage}
\vspace{-0.1in}
\end{table}

\begin{figure}[tp]
\centering
\begin{tabular}{c}
\includegraphics[scale=0.56]{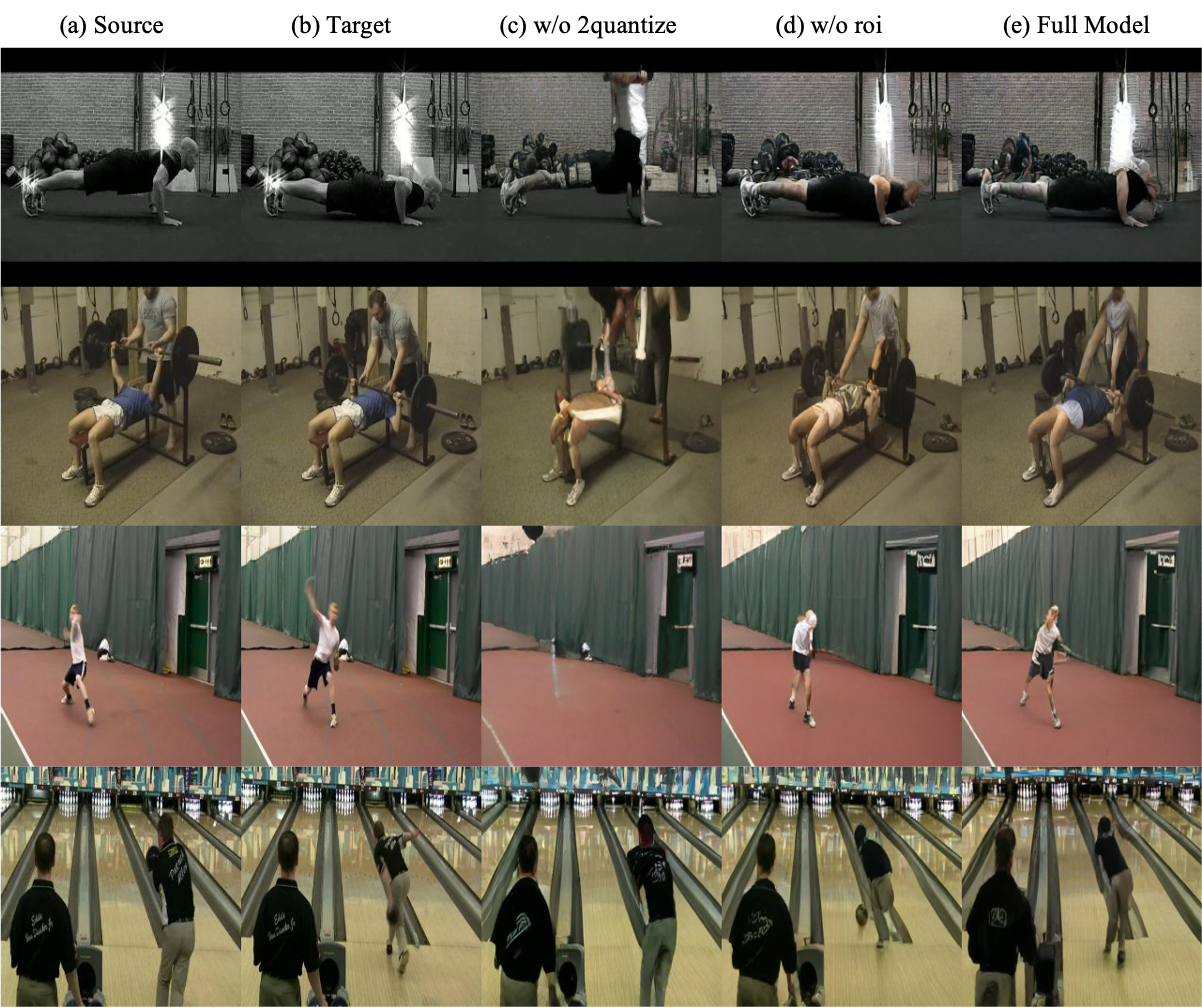}
\end{tabular}
 \vspace{-0.15in}
 \caption{Qualitative results in ablation study on PA dataset.
 \label{fig:pa_poseguided}}
\end{figure}

\subsection{Quantitative Results}
\label{subsec:exp-auantitative}
We also give the general measurement of the image quality. Thus we compare the 
quantitative results on PA dataset 
of our proposed QS-Craft and other competitors  in  Tab.~\ref{tab:pa_pose-guided}. As for Tai-Chi-HD dataset, since there is no ground truth according to the animation results to get quantitative scores, a sanity check by video reconstruction is conducted to demonstrate the effectiveness of our QS-Craft. Here we employ three metrics described above, Average Keypoint Distance (AKD), Missing Keypoint Rate (MKR) and Fr\'echet Inception Distance (FID) to measure the quality of generated results. 


Note that for Monkey-Net, FOMM, and MRAA are not designed to tackle the complex images from PA dataset; and thus there is no available model for the direct comparison. Thus, the methods of PG$^{2}$, PATN, PN-GAN and MR-Net are compared here. We note that our
QS-Craft outperforms all the baselines in MKR and FID, and ranks second in AKD. This intuitively reflects that our QS-Craft can deal with the complicated background with human motion,  producing visually good synthesized results.

\begin{figure}[tbp]
\begin{centering}
\includegraphics[scale=0.6]{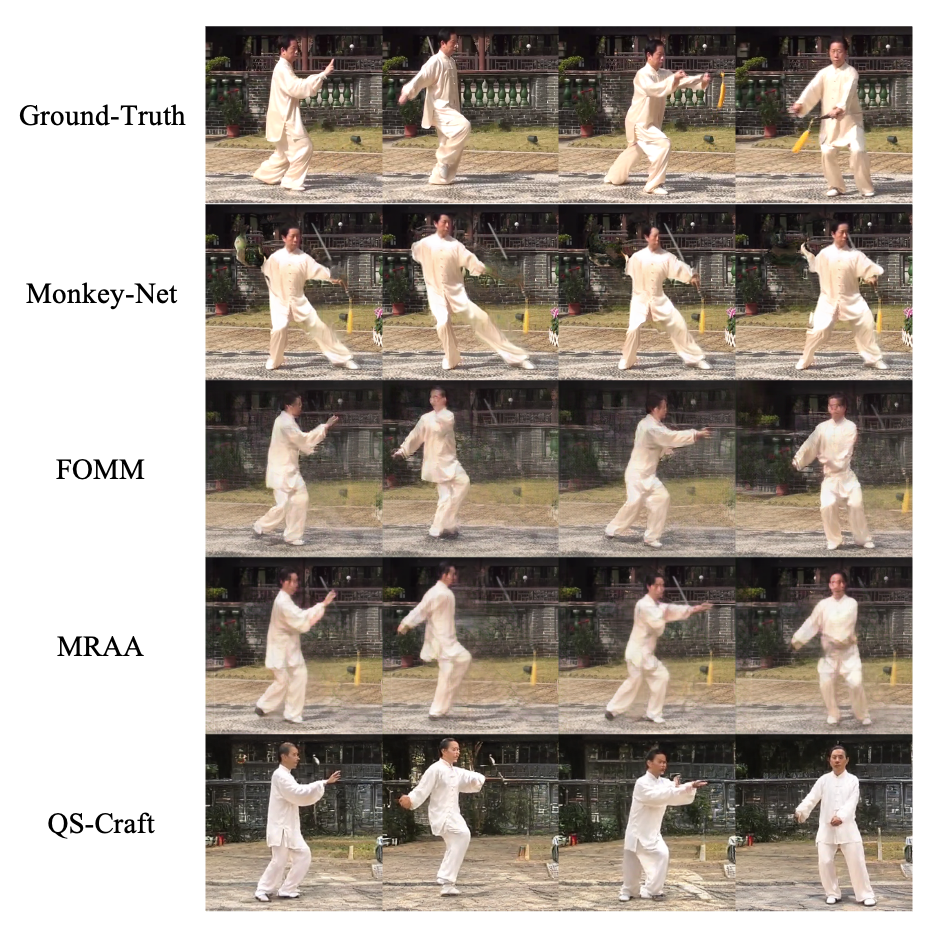}
\vspace{-0.15in}
\caption{Reconstruction results on Tai-Chi-HD. The synthesized results are compared among different methods. \label{fig:taichi_reconstruction}}
\end{centering}
 \vspace{-0.15in}
\end{figure}

\noindent \textbf{Sanity check by video reconstruction. }
We take this task as the sanity check of our model, as it is designed for animation in transferring settings. That is, source image and driving video are different.
In particular, we give results of video reconstruction in Fig.~\ref{fig:taichi_reconstruction}.
Image frames reconstructed by Monkey-Net suffer from the wrong motion compared against the ground truth. Furthermore, FOMM and MRAA can capture the general motion information whilst ignoring some body parts such as hand movements in the third column. 
In a word, our reconstructed images demonstrate the superiority to the quality and detailed texture, which highly supports the efficacy of our QS-Craft framework in addressing the cMHA task.
The quantitative results on Tai-Chi-HD dataset are shown in Tab.~\ref{tab:taichi_reconstruction}. We observe that our QS-Craft outperforms other methods in all metrics, which indicated the superiority of our method on animating accurate motions from driving videos as well as generating authentic and realistic foregrounds.

\noindent \textbf{User study. }
To complete our evaluation, a user study is conducted on Tai-Chi-HD dataset. We provide 32 users with two test sets of randomly selected animation results and ask them to select the most realistic and reasonable generation. Specifically, both test sets contain 25 different image animations, each of which involves one source image, one driving video and two animated generations. Results in Tab.~\ref{tab:taichi_user_study} show that our QS-Craft is preferred over the competitor.

\subsection{Ablation Study}
\label{subsec:ablation_pose}
In our framework, we particularly use pose coordinates, encoded by multilayer perceptrons, as conditional features. In order to verify the effectiveness of this design (termed \textit{Pose Coord}), we conduct three variants: (1) \textit{Pose Flow}: we follow \cite {siarohin2019animating} to build a motion net with five convolution layers to encode flow information. Differently, the flow features are further applied as a sequence in our transformer model, rather than warping source features \cite {siarohin2019animating}. (2) \textit{Pose Image}: Similar to other studies \cite {qian2018pose,xu2020pose,ma2017pose}, we use pose skeleton RGB images as condition information. Pose image features are extracted with five convolution layers. (3) \textit{Pose VQ}: We use pose images to train a VQ-VAE model \cite{oord2017neural} and then the quantized pose embeddings are applied as conditions, which is the same as \cite{esser2021taming}.

As illustrated in Fig.~\ref{fig:ablation}, the results achieved by our QS-Craft are apparently more natural than other variants. Particularly, our results are authentic with richer accurate motion information from the driving videos. 
Besides, quantitative results in Tab.~\ref{tab:ablation} also validate the same conclusion. For all the metrics, our \textit{Pose Coord} outperforms others apparently, which suggests the superiority of precisely guiding motion changes.

Furthermore, the second ablation study is conducted to demonstrate the effectiveness of our proposed \emph{Scrabble} step and RoI weight in Fig.~\ref{fig:pa_poseguided} and Tab.~\ref{table:pa_ablation}. As we can find in Fig.~\ref{fig:pa_poseguided}, model trained without \emph{Scrabble} step fails to generate reasonable target images with accurate pose information as it do not take into account the strong connection between the source and target in training phase and then accumulates prediction errors in inference time. Additionally, model lack of RoI weight pays less attention to foregrounds compared to our QS-Craft so it is inclined to generate blurred human motion or different clothes colors. And quantitative results of our QS-Craft in Tab.~\ref{table:pa_ablation} outperform other variants in AKD and FID, which indicates that our full model can generate more convincing and photo-realistic images with accurate target pose conditions.

\section{Conclusion}
In this paper, we propose a novel method to animate objects conditioned on driving videos through three phases: \emph{Quantize}, \emph{Scrabble} and \emph{Craft}. This transformer-based model can effectively generate semantically consistent and realistic results as we demonstrate above. Besides, especially compared to other methods like Monkey-Net, FOMM and MRAA which aim to handle the animation task, our proposed QS-Craft do not need the source image given to be aligned with the first frame of the driving video as the simple pose keypoints are enough for QS-Craft. 
Our QS-Craft also encounters some challenges, including smoothness in video animation, since there is no temporal information involved in our method.

\clearpage
%
%
\bibliographystyle{splncs04}
\bibliography{egbib}
\end{document}